\documentclass[10pt,twocolumn,letterpaper]{article}
\usepackage{booktabs}
\usepackage{array}
\usepackage{multirow}
\usepackage{amsmath}
\usepackage{graphicx}
\usepackage{subcaption}
\usepackage{stfloats}
\usepackage{xcolor} 

\makeatletter

\usepackage{cvpr}      
\usepackage[T1]{fontenc}
\usepackage{subcaption}
\usepackage{makecell}

\usepackage[dvipsnames]{xcolor}

\definecolor{cvprblue}{rgb}{0.21,0.49,0.74}
\usepackage[pagebackref,breaklinks,colorlinks,citecolor=cvprblue]{hyperref}

\def\confName{3DV\xspace}
\def\confYear{2025\xspace}

\title{No Mesh, No Problem: Estimating Coral Volume and Surface from Sparse Multi-View Images}

\author{
Diego E. Farchione \quad Ramzi Idoughi \quad Peter Wonka \\
KAUST, Saudi Arabia \\
{\tt\small \{diegoeustachio.farchione,ramzi.idoughi,peter.wonka\}@kaust.edu.sa}
}

\definecolor{gray}{gray}{0.6}

\definecolor{darkblue}{rgb}{0,0,0.9}
\definecolor{blue}{rgb}{0,0,1}
\definecolor{orange}{rgb}{1,.5,0} 
\definecolor{red}{rgb}{1,0,0} 

\newcommand{\Volume}[1]{{\color{orange} #1}}
\newcommand{\Surface}[1]{{\color{blue} #1}}

\renewcommand{\etal}{\text{et~al.~}}

\newcommand{\Figure}{Figure~}

\newcommand{\Section}{Section~}

\newcommand{\Loss}[1] {\mathcal{L}_{\text{#1}}}

\newcommand{\mean}{\ensuremath{\mu}}  
\newcommand{\std}{\ensuremath{\sigma}}

\begin{document}
\maketitle

\begin{abstract}
Effective reef monitoring requires the quantification of coral growth via accurate volumetric and surface area estimates, which is a challenging task due to the complex morphology of corals. We propose a novel, lightweight, and scalable learning framework that addresses this challenge by predicting the 3D volume and surface area of coral-like objects from 2D multi-view RGB images. Our approach utilizes a pre-trained module (VGGT) to extract dense point maps from each view; these maps are merged into a unified point cloud and enriched with per‑view confidence scores. The resulting cloud is fed to two parallel DGCNN decoder heads, which jointly output the volume and the surface area of the coral, as well as their corresponding confidence estimate. To enhance prediction stability and provide uncertainty estimates, we introduce a composite loss function based on Gaussian negative log-likelihood in both real and log domains. Our method achieves competitive accuracy and generalizes well to unseen morphologies. This framework paves the way for efficient and scalable coral geometry estimation directly from a sparse set of images, with potential applications in coral growth analysis and reef monitoring.

\end{abstract}  
\section{Introduction}
\label{sec:intro}

Coral reefs are an important component of marine life and serve as a vital indicator of the health of marine ecosystems. Tracking metrics like coral volume and surface area can play a crucial role in understanding growth trends, detecting degradation over time, and guiding conservation efforts. Yet, the traditional methods for estimating these coral features, relying on manual measurements or 3D reconstruction pipelines~\cite{koch20213d, lange2020quick,yudelman2022coral,helmholz2024evaluating}, require high-end scanning equipment, heavy human involvement, and laborious post-processing. Such approaches are not scalable, which limits their practical utility in large-scale monitoring initiatives. Furthermore, when complete point clouds are unavailable, for instance, due to limited image capture or occluded coral regions, conventional mesh reconstruction methods like Poisson surface reconstruction~\cite{Kazhdan2006PoissonSurfaceReconstruction} often fail to generate accurate meshes. This incomplete geometry can introduce artifacts (e.g., planes, voids, and extruding bump-like surfaces) that can severely compromise the reliability of volume and surface estimates.

Recent advances in computer vision and deep learning offer promising alternatives by enabling 3D understanding tasks to be performed directly from 2D imagery. This includes tasks such as monocular depth estimation, pose estimation, and deformation estimation. Notably, multi-view techniques have shown success in estimating geometry without requiring full 3D supervision ~\cite{tulsiani2017multi,ranftl2020towards,kanazawa2018end, godard2019}. However, adapting such frameworks to complex, porous coral structures poses significant challenges: extreme morphological variability, high frequency of fine geometric details, and the lack of large-scale annotated datasets.

In this work, we propose a scalable and lightweight learning pipeline (\Figure~\ref{fig:fullwidth_coral}) that estimates jointly coral volume and surface area directly from 2D multi-view RGB images. Our framework leverages the Visual Geometry Grounded Transformer (VGGT)~\cite{wang2025vggt}, a vision-language model tailored for geometry-related tasks, to extract dense point maps from masked coral images captured across various viewpoints. These multi-view point maps are then converted into a 3D point cloud. The main advantage of using VGGT in our application is its ability to produce high-quality, dense point clouds, regardless of the number of available images, making it effective even with a sparse set of 2D images. The resulting point clouds are augmented with their corresponding confidence scores output by VGGT, and subsequently processed by two decoder heads based on the Dynamic Graph Convolutional Neural Network (DGCNN)~\cite{wang2019dynamicgraphcnnlearning}. The first decoder head regresses the coral volume, while the other targets the coral surface area, with each head also generating a prediction confidence score, enabling users to assess the uncertainty associated with the estimates.
To deal with the inherently high uncertainty of this problem, we design a hybrid loss function that combines Gaussian Negative Log-Likelihood (NLL) ~\cite{lakshminarayanan2017simple} in both linear and logarithmic spaces with deterministic metrics, including Mean Absolute Error (MAE), log-MAE, and Relative Error. This combination promotes model robustness across diverse coral sizes and shapes while producing meaningful uncertainty estimates.

Our method achieves competitive results and generalizes effectively to various coral morphologies. This approach opens the door to fast, automated image-based assessments of coral structure and metrics. Therefore, it has significant potential for large-scale reef conservation efforts.\\

\noindent The \textbf{key contributions} of this work are as follows:\\
\begin{itemize}
    \item We propose a lightweight, scalable, and fully automated pipeline for estimating coral volume and surface area from 2D multi-view RGB images, without requiring explicit 3D mesh supervision during inference.
    
    \item We harness VGGT to generate multi-view point maps from masked coral images and employ two DGCNN-based decoders to predict geometric properties from aggregated, possibly incomplete, 3D point clouds.
    
    \item We design a hybrid loss function that combines Gaussian NLL in linear and logarithmic domains with deterministic error terms (e.g., MAE, log-MAE, and relative error), fostering uncertainty-aware training and stability across different coral scales.
    
    \item Our framework mitigates artifacts induced by traditional mesh-reconstruction methods, enabling accurate predictions even with limited viewpoints and partial visibility.
    
    \item We demonstrate the effectiveness of our method on various synthetic coral meshes, achieving a low mean absolute percentage error despite high structural complexity and limited training data.
\end{itemize}

\section{Related Work}
\label{sec:related}

\subsection{3D Reconstruction in Computer Vision}
\label{sec:3d_reconstruction}

3D reconstruction is a fundamental challenge in computer vision, aiming to recover the 3D structure of a scene or object from 2D captured images. Classical approaches rely on Structure-from-Motion (SfM) and Multi-View Stereo (MVS) pipelines, such as COLMAP~\cite{Schonberger_2016_CVPR,Schonberger_2016_ECCV_MVS}, which estimate camera poses and dense point clouds from image sequences. These methods have demonstrated strong performance across diverse application domains. However, they typically require extensive preprocessing, high-quality feature correspondences, and fail under sparse viewpoints or challenging conditions (e.g., textureless regions, dynamic lighting), which is the case in underwater environments.

Learning-based techniques have addressed these limitations by incorporating deep neural networks into the reconstruction pipelines. For instance, MVSNet~\cite{yao2018mvsnet} introduced an end-to-end depth estimation through cost volume regularization, while CasMVSNet~\cite{gu2020casmvsnet} and PatchMatchNet~\cite{wang2021patchmatchnet} extended this framework with hierarchical refinement and learned patch matching to enhance accuracy and robustness in unconstrained environments.

More recent advancements have shifted toward implicit neural representations, initially introduced by Neural Radiance Fields (NeRFs)~\cite{mildenhall2020nerf} for the photorealistic novel view synthesis task. Extensions such as PlenOctrees~\cite{yu2021plenoctrees}, FastNeRF~\cite{yu2021fastnerf}, Mip-NeRF~\cite{barron2021mipnerf}, and Instant-NGP~\cite{mueller2022instant} improve efficiency. Another branch of learning-based approaches has been started with 3D Gaussian Splatting~\cite{oreshkin2023gaussian}, which performs real-time rendering through Gaussian primitives. Nevertheless, these extensions still require dense views and accurate camera parameters. By leveraging a structured 3D latent representation Trellis~\cite{xiang2024trellis} generates a high-quality 3D asset from a single image and a text prompt. VGGT~\cite{wang2025vggt} introduces a transformer-based framework for recovering consistent geometry from unposed unordered image sets, pushing the boundary of unstructured 3D reconstruction.

\subsection{Deep learning techniques for Coral Monitoring}

Deep learning has emerged as a powerful tool for coral reef monitoring, with applications primarily focused on \textit{coverage estimation}, \textit{segmentation}, and  \textit{species identification}. For instance, attention-based segmentation models such as PSPNet have been used to estimate live coral cover from underwater video, achieving high accuracy in reef mapping~\cite{li2024deep}. CoralSCOP~\cite{zheng2024coralscop} has been proposed as a foundation model for producing dense segmentation of coral reefs. Other studies have focused on species-level classification in complex underwater imagery~\cite{Raphael2020,sauder2025coralscapes}, facilitating biodiversity assessment. 

Beyond 2D analysis, 3D reconstruction has also been used to monitor reef structural complexity. Hopkinson \etal\cite{Hopkinson2020} combined SfM pipelines with CNN to produce labeled 3D reef meshes. More recent advancements leverage drones and stereo or multi-view imagery for large-scale mapping of reef rugosity and volumetric structures. Notably, Suan \etal\cite{SUAN2025102958} proposed a deep-learning-based drone pipeline that predicts 3D structural complexity metrics like rugosity from aerial color images. However, these approaches typically retrieve the area coverage metrics or topographic indices rather than direct coral volume estimation.

\subsection{Neural Volume Estimation}
Neural networks have demonstrated significant potential for accurate volume estimation across various domains. For example, VolNet~\cite{leinen2021volnet} predicts human body volume from a single RGB image by combining segmentation with regression, achieving high accuracy for anthropometric applications like health and fitness assessments. Similarly, in the food domain, deep learning methods have been used to estimate meal volumes from depth maps or single/few RGB-D images~\cite{fang2016,Fang2018,VolETA2024}. These approaches often integrate view synthesis, point cloud completion, or generative mesh reconstruction to handle complex shapes and occlusions. Industrial applications have also benefited from neural volume estimation. Kim \etal\cite{Kim2025} proposed a patch-based neural approach that processes paired color and depth images to estimate the volume of irregular objects, offering robust solutions for quality control and manufacturing processes. Despite the strong performance of these approaches, to the best of our knowledge, no prior work has directly utilized neural methods to estimate the 3D volume of individual coral colonies. This research gap motivates our approach, which combines 3D reconstruction and neural regression, enabling precise and scalable measurement of coral volumes from images.


\begin{figure*}[t]
    \centering
    \includegraphics[width=0.98\textwidth,height=0.312\textheight]{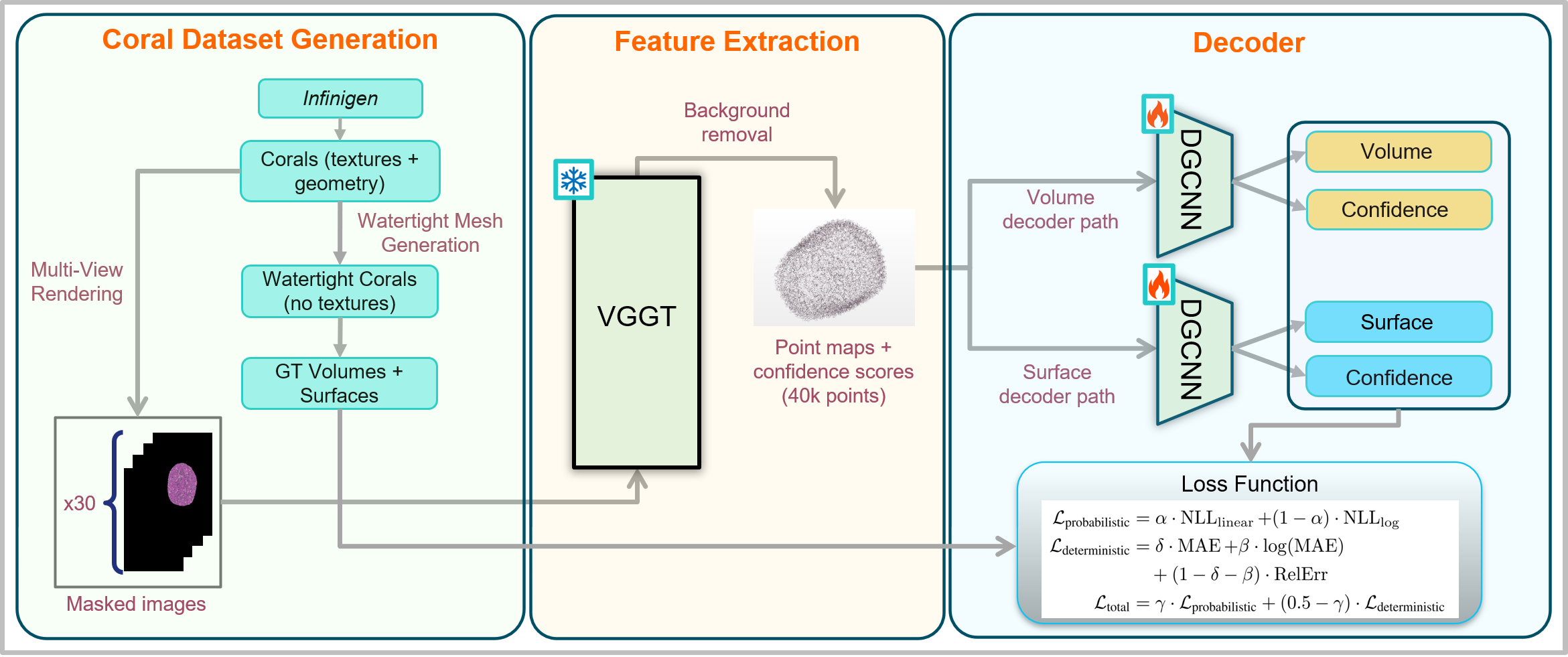}
    \caption{Overall architecture of the proposed pipeline for coral volume and surface estimation. 
The \textbf{Coral Dataset Generation} block produces coral meshes using \textit{Infinigen}, then converts them to watertight meshes, and computes ground-truth (GT) volumes and surface areas. Multi-view rendering is carried out using a black background to obtain the input images.
In the \textbf{Feature Extraction} block, the pointmap branch of VGGT takes masked images as input and outputs dense point maps along with confidence scores. 
A background removal step is applied to eliminate irrelevant background points, and the filtered point maps are converted into 3D point clouds. 
In the \textbf{Decoder} component, two DGCNN-based decoder paths regress the coral’s surface area and volume, each with an associated confidence score. 
The network is trained using a combination of deterministic and probabilistic losses.}

    \label{fig:fullwidth_coral}
\end{figure*}  

\section{Method}
\label{sec:method}

\subsection{Overview}
\label{ssec:overview}
The architecture of our approach, shown in \Figure \ref{fig:fullwidth_coral}, has three main components: Coral dataset generation, feature extraction, and decoder. In the first step, we synthesize watertight coral meshes to obtain precise ground-truth volumes and surface areas. We also render sparse multi-view RGB images from varied, known camera poses. The second block of our framework applies the pre-trained VGGT pointmap branch to process these multi-view images to produce dense, confidence-weighted point maps that are converted into 3D point clouds. Finally, two parallel DGCNN decoder heads independently predict the coral volume and surface area and output an associated confidence score for each estimate. In the following, we describe each component of our framework in detail.

\subsection{Coral Dataset Preparation}
\label{ssec:dataset_generation}

\subsubsection{Synthesis of Watertight Coral Meshes and Output Metrics Computation.}
\begin{figure}[!ht]
\centering
\begin{subfigure}{0.48\linewidth}
  \includegraphics[width=\linewidth]{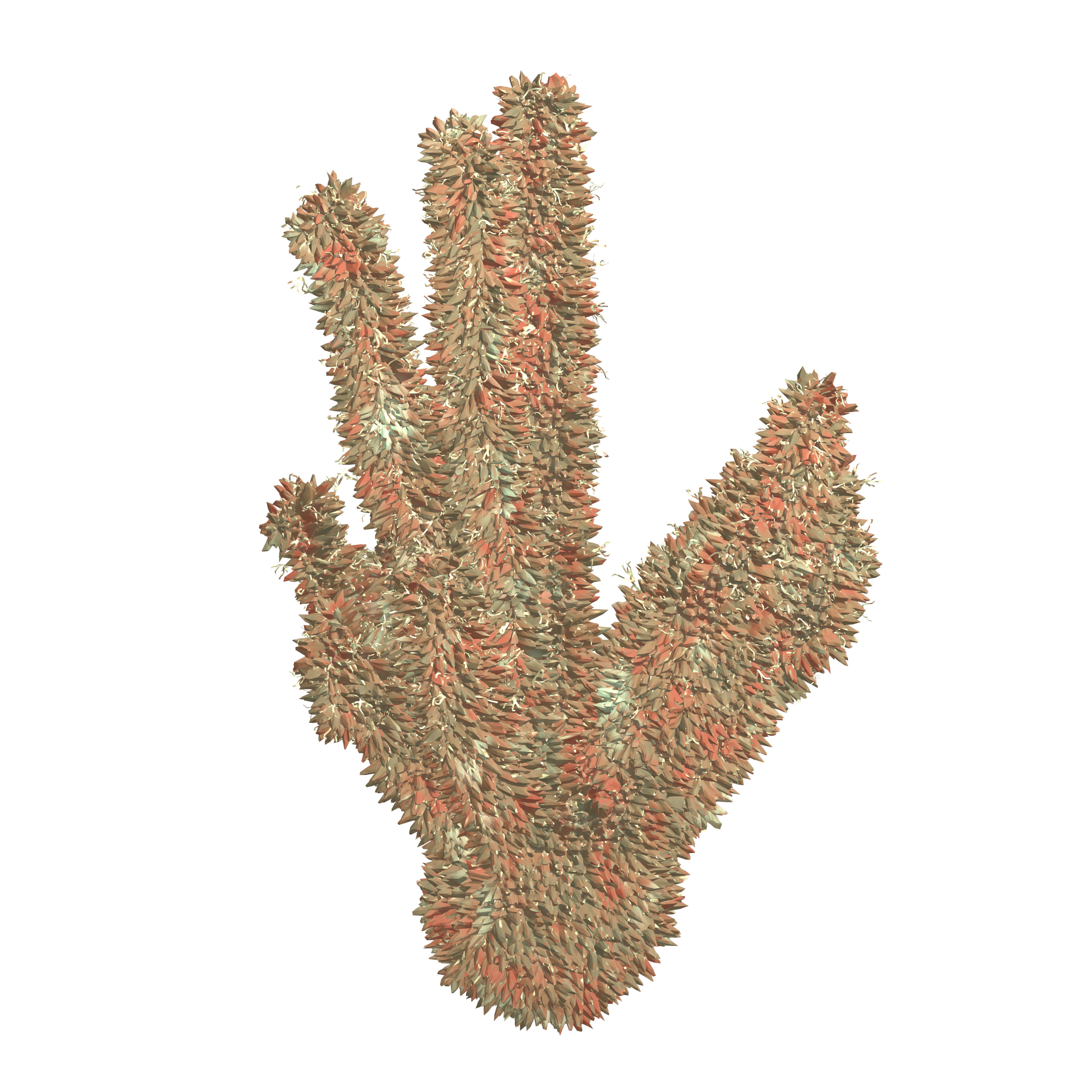}
\end{subfigure}
\begin{subfigure}{0.48\linewidth}
  \includegraphics[width=\linewidth]{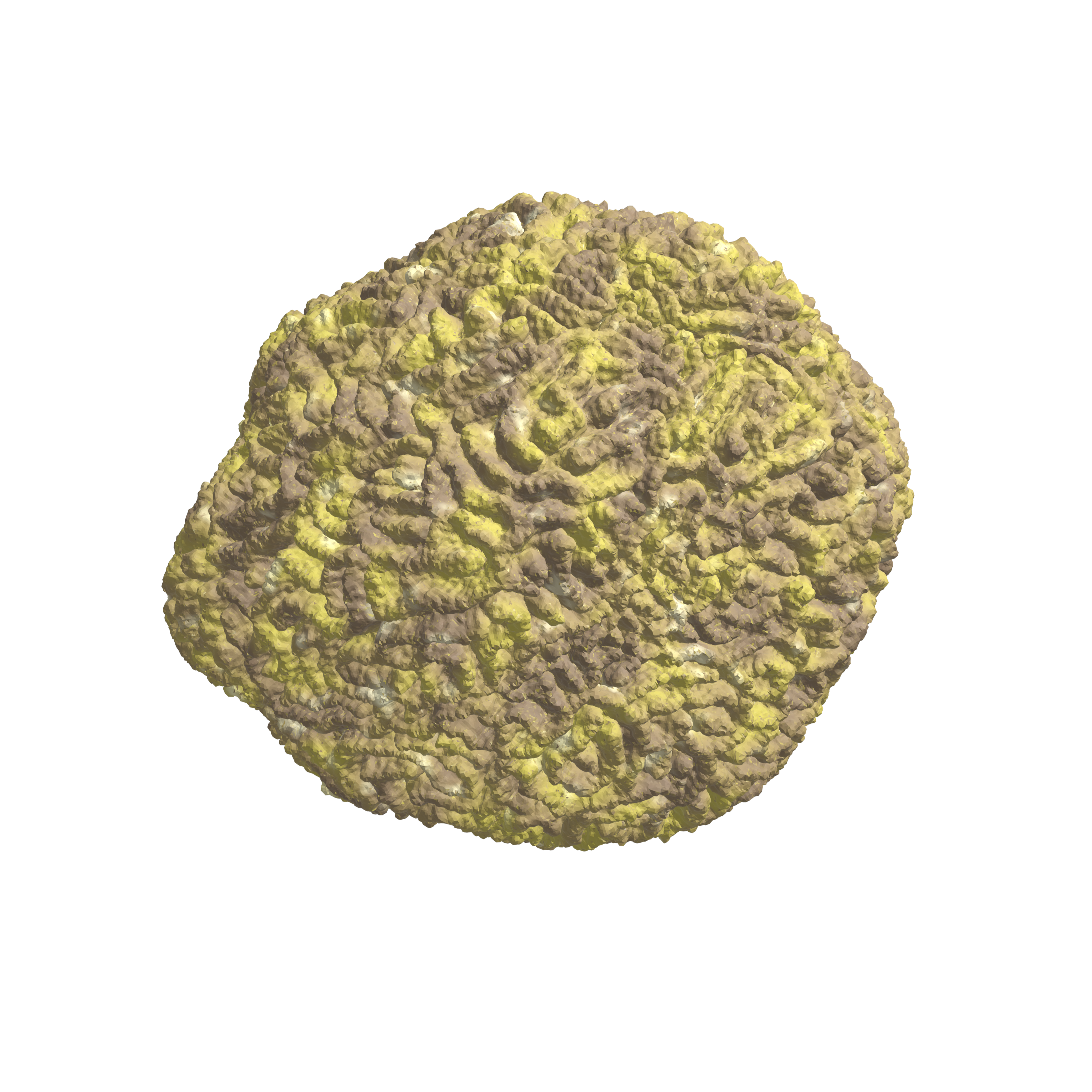}
\end{subfigure}


\begin{subfigure}{0.48\linewidth}
  \includegraphics[width=\linewidth]{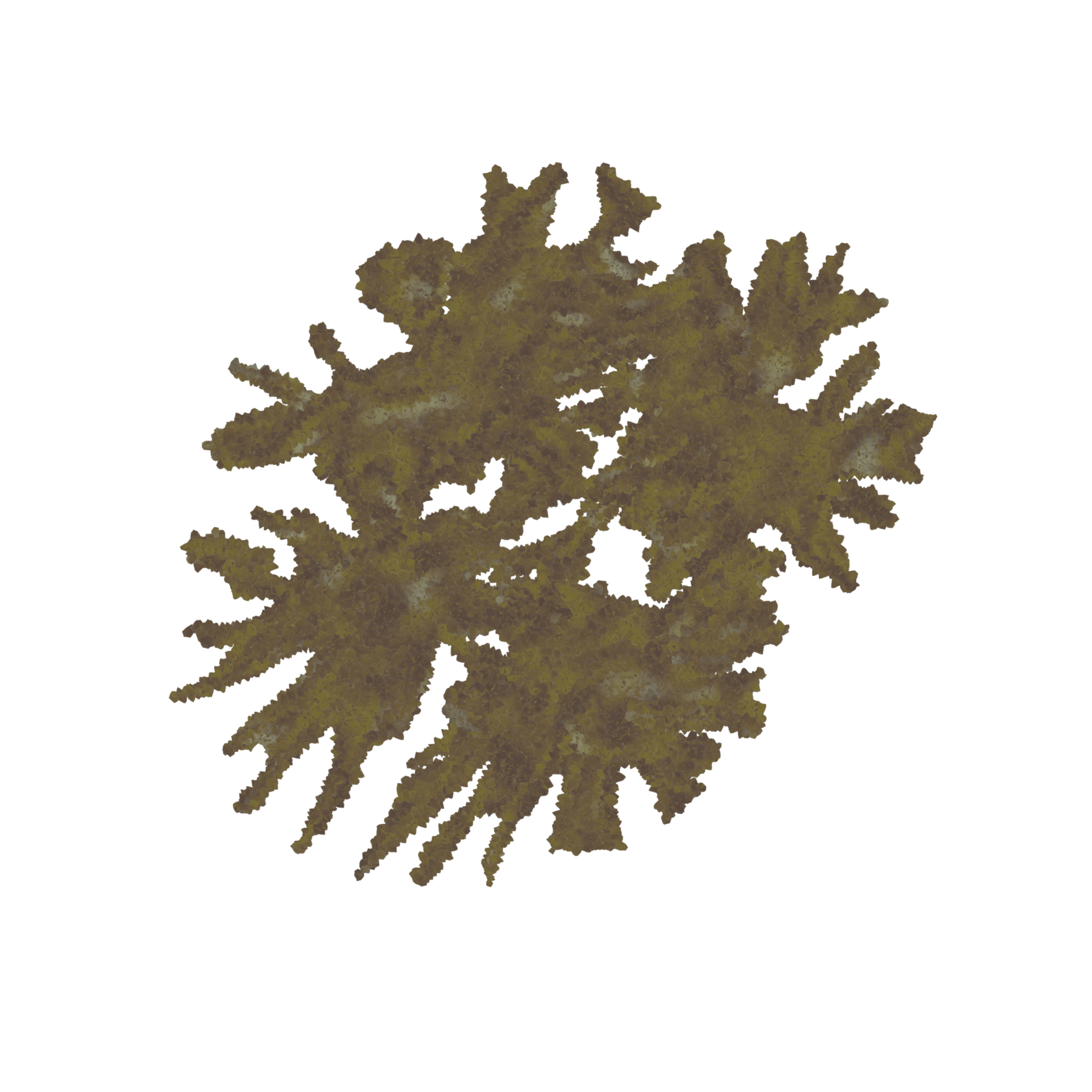}
\end{subfigure}
\begin{subfigure}{0.48\linewidth}
  \includegraphics[width=\linewidth]{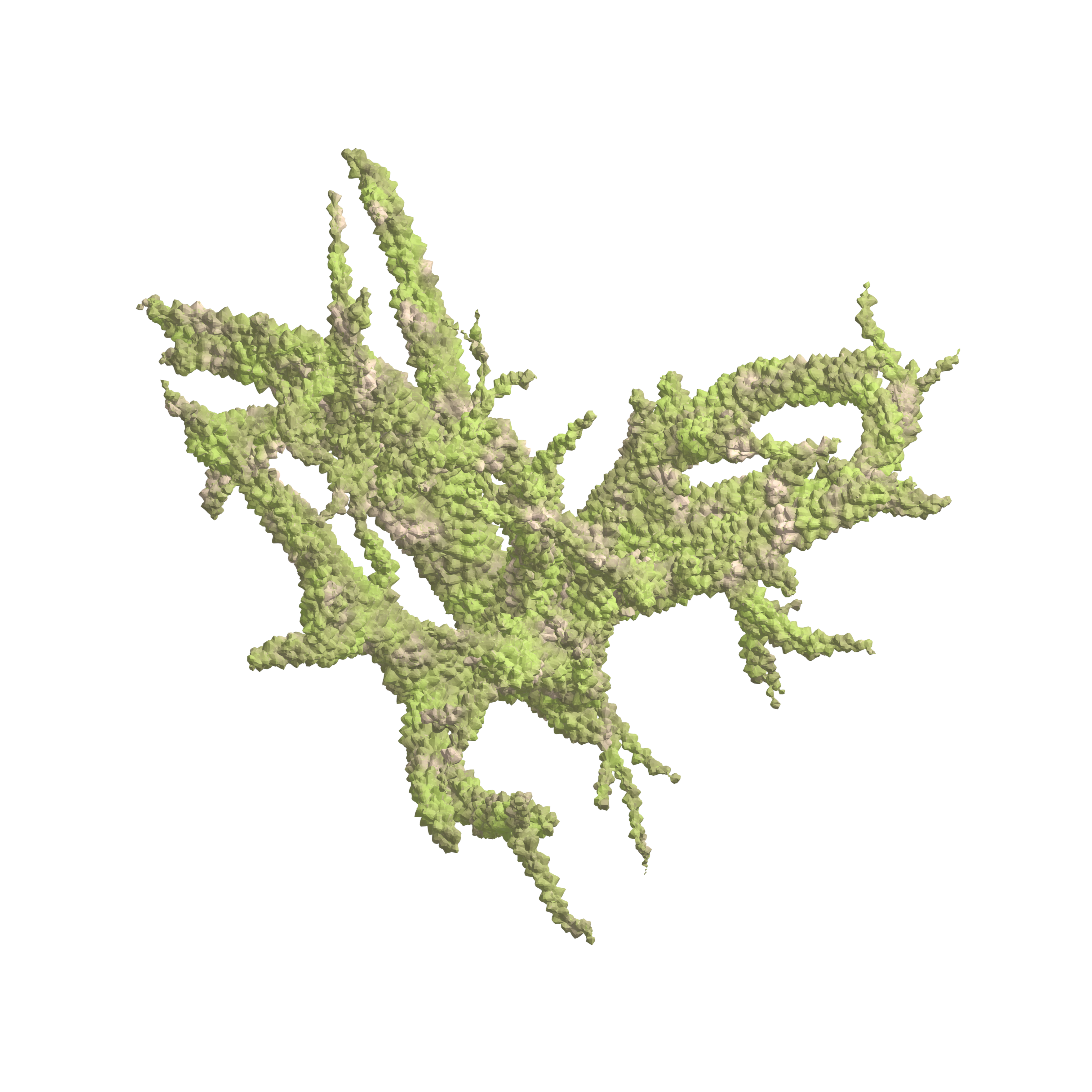}
\end{subfigure}
\caption{Examples of 4 different coral morphologies. The generated corals exhibit highly complex geometries, including concave regions, surface cavities, irregular contours, and intricate patterns.}
\label{fig:coral_examples}
\end{figure}

To train our proposed network to accurately estimate coral volumes, we generate watertight 3D meshes from coral models. Given the absence of readily available real-world coral data, this synthetic generation is crucial for providing diverse and reliable training samples. \Figure\ref{fig:coral_examples} illustrates 4 synthetically generated corals. The process begins with assets obtained 
from \textit{Infinigen}~\cite{raistrick2023infinite}, and are further processed using \textit{ManifoldPlus}~\cite{huang2020manifoldplus} to obtain watertight meshes of corals. This step is crucial for an accurate geometric computation of ground-truth volumes and surface areas.
Following the watertight transformation, the required ground-truth metrics of the coral (volume and surface area) are computed from those assets
using PyVista~\cite{sullivan2019pyvista} and Trimesh~\cite{trimesh} (refer to the Supplementary Section for details).

To promote stability, efficiency, and good generalization of our training process, we propose using normalized volumes and surface areas. Specifically, before computing the ground-truth metrics of each coral, each mesh is normalized by scaling its bounding box such that its maximum dimension is equal to one.



\subsubsection{Input Images Rendering and Masking.}
The input of our framework is a sparse set of captured images of the coral from different viewing angles. We generate such entries for the synthetic dataset by rendering RGB images using PyVista on the textured 
coral before applying \textit{ManifoldPlus}, which removes the texture and outputs only the watertight geometry. Each image has a resolution of $1024\times 768$ pixels, with the camera's field of view ranging from $22^{\circ}$ to $40^{\circ}$. A random slight jitter is applied to each camera position to prevent it from pointing directly at the object's center.
Backgrounds are removed during rendering, resulting in black-padded images centered on the coral geometry, with a small random translation applied.
To simulate realistic lighting conditions, we manually added 32 scene lights. These lights are evenly distributed in azimuth over $360^{\circ}$, with elevations randomly set between $10^{\circ}$ and $80^{\circ}$. Each light has an intensity of $0.18$.
The mesh is rendered with texture mapping and physically-based shading parameters: an ambient coefficient of $0.2$, a diffuse coefficient of $0.9$, a specular coefficient of $0.4$, and full opacity. This configuration enhances surface details through multi-directional illumination and aligns with underwater environments, where multi-path scattering is equivalent to multiple light sources.

\subsubsection{Error Mitigation and Dataset Filtering.}
During the dataset generation process, errors primarily arise from the watertight mesh conversion step using \textit{ManifoldPlus}. These include small inaccuracies in surface reconstruction, such as the creation of thin, disconnected surfaces or artifacts in regions with sparse or noisy geometry, which can significantly alter the computed volume and surface area, especially for smaller corals. Additionally, for objects with very small volumes, even minor geometric errors during watertight closure can lead to substantial relative changes, as the algorithm may force closures in isolated or poorly defined areas, deviating from the original coral.

To mitigate these issues and ensure a high-fidelity dataset, we performed a post-processing filtering step after the first part of the training. Specifically, we rejected samples with the worst $3\%$ of predictions in terms of absolute or relative errors. In the supplementary, are shown examples of faulty watertight reconstructions. This filtering step helps maintain dataset quality by focusing on meshes that accurately represent the original coral geometry, thereby improving the reliability of the ground-truth metrics and the training stability.

\subsection{Feature map extraction using VGGT}
\label{ssec:VGGT}
Our processing pipeline begins by generating dense pointmaps from masked coral images using the pointmap branch of the pre-trained VGGT framework. This step infers a point map for each input image, enriched with a corresponding confidence score map to quantify the model's reliability in the reconstruction process. These point maps are merged together to build a consistent point cloud. To refine the point cloud and eliminate background noise, we apply two key filters from the VGGT framework: the black background filter, which prevents the generation of points from non-coral areas, and a learning-based background removal filter, which we found significantly improves results by removing irrelevant background elements.
Finally, we concatenate the filtered point cloud with the associated confidence scores to create a \textbf{feature map} that represents the morphology of the coral. This representation offers substantial flexibility to select the number of points to be used as input to the decoder heads.

\subsection{DGCNN-based Decoders}
\label{ssec:decoder}
The extracted feature maps are fed into two parallel DGCNN models, which are trained to predict the coral metrics and their corresponding confidence scores. We employ two separate decoders for volume and surface area predictions to enable the network to independently optimize feature extraction for each metric. By exploiting spatial relationships between points, this model effectively captures both local intricacies and global geometric features of the corals. We selected this decoder architecture because it delivers superior performance for volume estimation and a balanced trade-off for surface estimation, as detailed in \Section\ref{ssec:architecture_analysis}. Implementation details on the DGCNN decoders' architecture can be found in the supplementary material.

\subsection{Loss Function}
\label{ssec:loss}
In our training process, we adopt a hybrid loss function that combines probabilistic and deterministic components. On the probabilistic side, our loss integrates linear and logarithmic Gaussian Negative Log-Likelihood (NLL) terms to handle uncertainty, while incorporating error metrics for robust prediction accuracy. This design is inspired by the work of Lakshminarayanan \etal\cite{lakshminarayanan2017simple}, where the Gaussian NLL assumes the target variable to follow a normal distribution with mean $\mean$ (the predicted value) and a learned variance $\std^2$. The Gaussian NLL term encourages the model not only to produce accurate predictions but also to estimate the associated uncertainty. We use a weighted combination of the NLL in the original (linear) space and in the logarithmic space of the target to accommodate targets with a wide range of magnitudes, which is the case for both coral metrics. The Gaussian NLL in the linear space is given by the following equation:

\begin{equation}
\operatorname{NLL_{\mathrm{linear}}}(y, \mean, \log \std^2) = \frac{1}{2} \left( \log \std^2 + \frac{(y - \mean)^2}{\std^2} \right)
\label{eq:NLL}
\end{equation}

where $y$ is the ground truth metric, $\mean$ and $\std^2$ are respectively the mean and the variance of the predicted metric. The logarithmic Gaussian NLL ($\operatorname{NLL_{\log}}$) applies the equation~\ref{eq:NLL} to the logarithmically transformed data.

On the deterministic side, the loss includes three key terms to penalize prediction errors: (1) The \textbf{Mean Absolute Error (MAE)} penalizes the absolute deviation between prediction and ground truth, providing a robust and interpretable measure of error. (2) The \textbf{Log-MAE} reduces sensitivity to large outliers by applying a logarithmic transformation to MAE. (3) \textbf{Relative Error} is defined as $\operatorname{RelErr} = \frac{|y - \hat{y}|}{y}$ and allows the model to account for proportionally larger errors when the target values are small.

By combining these probabilistic and deterministic elements, the loss function enables the model to minimize prediction errors while simultaneously estimating uncertainty, effectively addressing data variability in volume and surface area predictions. The total loss is formulated as follows:

\begin{align}
\Loss{probabilistic} &= \alpha \cdot \operatorname{NLL}_{\mathrm{linear}} + (1 - \alpha) \cdot \operatorname{NLL}_{\log} \nonumber \\
\Loss{deterministic} &= \delta \cdot \operatorname{MAE} + \beta \cdot \log(\operatorname{MAE}) \nonumber \\
                    &\quad + (1 - \delta - \beta) \cdot \operatorname{RelErr} \nonumber \\
\Loss{total} &= \gamma \cdot \Loss{probabilistic} + (0.5 - \gamma) \cdot \Loss{deterministic}
\label{eq:total_loss_eq}
\end{align}

where $\alpha$, $\beta$, $\delta$, and $\gamma$ are weights associated with the different loss terms.

\section{Results and Discussion}
\label{sec:results}

\subsection{Experimental Settings}

\noindent \textbf{Implementation Details.} We implemented the entire pipeline using Python and PyTorch libraries. Key libraries include PyTorch Geometric for testing some well-known models as decoders and PyVista/Trimesh for ground-truth computation and rendering, as already mentioned. All experiments were run on a single Nvidia A100 GPU on a Linux cluster. For the training step, we used a dataset of $1170$ training and $291$ validation corals. After applying VGGT to extract the feature map, we subsampled the point clouds to $40,000$ points featuring 4D attributes (3D position $+$ confidence score). The DGCNN architecture includes five EdgeConv blocks, global max pooling, and an MLP for volume/surface area prediction with uncertainty estimation. We employed the AdamW optimizer (learning rate: $1.6 \times 10^{-4}$, weight decay: $1.0 \times 10^{-4}$) with Cosine Annealing scheduling, training for up to 650 epochs and a batch size of 1, with an early stopping based on the validation loss. More details are provided in the Supplement.\\

\noindent \textbf{Evaluation metrics.} Since the corals in our dataset vary widely in size, absolute errors such as MAE and Root Mean Square Error (RMSE) are not sufficient to capture relative prediction quality. A small coral and a large coral can have the same absolute error, but the impact on accuracy is very different.
For example, an MAE of $0.02~\mathrm{m}^3$ is negligible for a coral of $2~\mathrm{m}^3$ but represents a $20\%$ error for a coral of $0.1~\mathrm{m}^3$.
Therefore, we report the Mean Absolute Percentage Error (MAPE) as our primary evaluation metric for volume prediction, as it normalizes the error by the true volume, providing a scale-independent measure that better reflects performance across corals of different sizes.

\subsection{Comparison with Trellis}
\label{sec:trellis_comparison}

To contextualize the performance of our method, we compare it against \textbf{Trellis}~\cite{xiang2024trellis}, which can generate high-quality 3D assets from a single RGB image. Although Trellis is not specifically designed for volume estimation, it offers an interesting baseline due to its ability to produce watertight meshes directly from images. These predicted meshes are then used to compute volumes and surface areas.

We evaluated Trellis on 15 coral samples from our dataset. For each sample, we provided 30 masked RGB images rendered from evenly spaced viewpoints around the coral, replicating the setup of our framework. Trellis processed each image independently to generate a 3D mesh, from which the volume and surface area were computed using standard mesh processing techniques. The final prediction for each coral was obtained by averaging the 30 individual predictions for volume and surface area. Table~\ref{tab:trellis_comparison} presents the mean MAPE obtained over the 15 corals for both the volume and surface area predictions. One can find the other evaluation metrics in the Supplementary Section.

\begin{table}[!h]
\centering
\caption{Comparison between our method and Trellis.}
\label{tab:trellis_comparison}
\scriptsize
\begin{tabular}{lccc}
\toprule
\textbf{Method}  $\downarrow$ & \textbf{Vol. MAPE} $\downarrow$ & \textbf{Surf. MAPE} $\downarrow$ \\
\midrule
Trellis &  60.81\% & 55.41\% \\
Ours (DGCNN)  & \textbf{10.27\%} & \textbf{7.56\%} \\
\bottomrule
\end{tabular}
\end{table}

These results shows that our model significantly outperforms Trellis in both volume and surface area estimation. This performance advantage arises primarily because our framework is specifically tailored to these tasks, even though it is based on point clouds rather than meshes.


\subsection{Impact of Training Dataset Size}

\begin{figure*}[!t]
    \centering
    \begin{subfigure}[b]{0.32\textwidth}
        \centering
        \includegraphics[width=\textwidth,height=0.22\textheight]{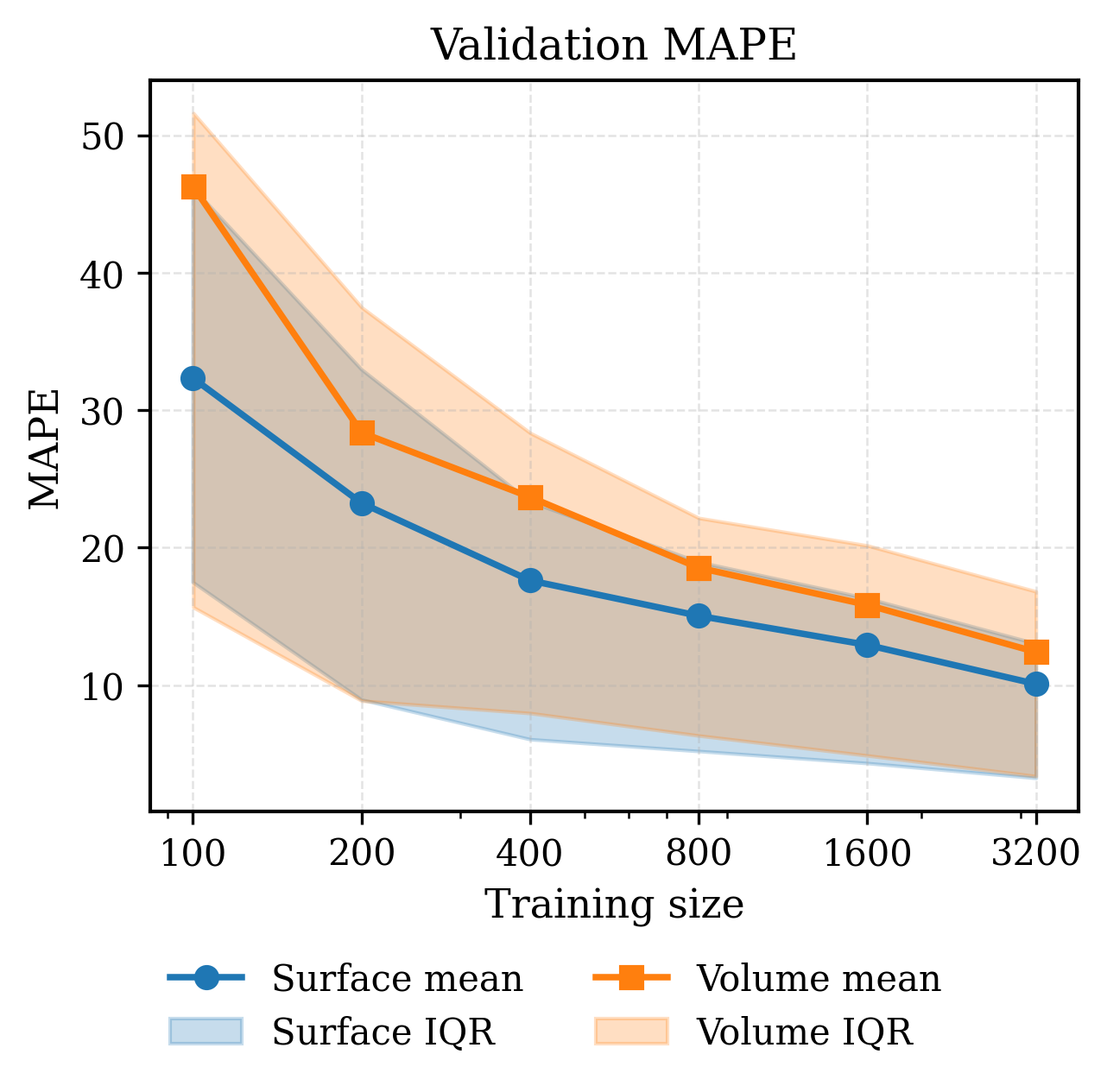}
        \caption{Validation MAPE (Volume \& Surface)}
        \label{fig:validation_mape_confidence}
    \end{subfigure}
    \hfill
    \begin{subfigure}[b]{0.32\textwidth}
        \centering
        \includegraphics[width=\textwidth,height=0.22\textheight]{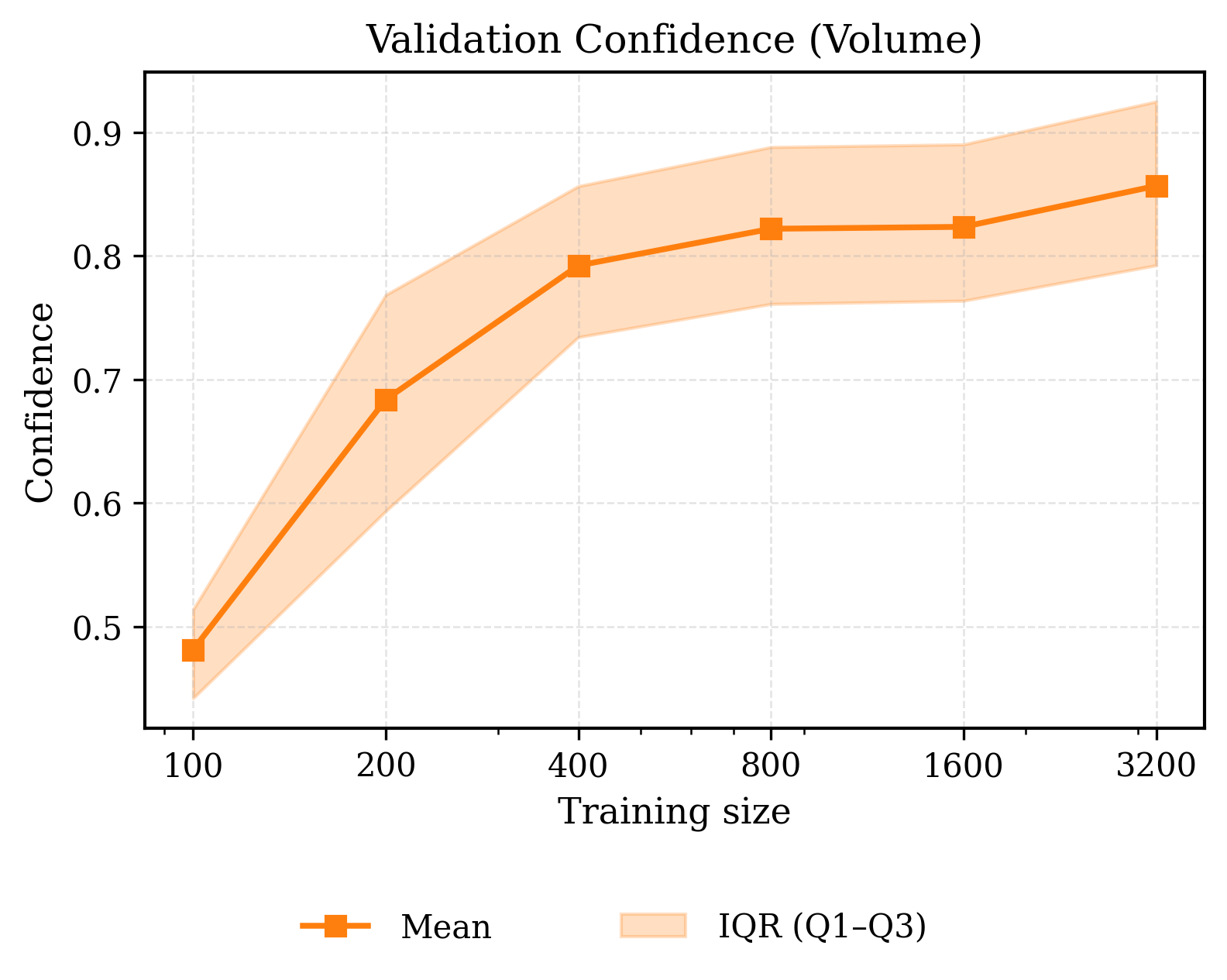}
        \caption{Confidence for Volume Estimation}
        \label{fig:confidence_volume_iqr}
    \end{subfigure}
    \hfill
    \begin{subfigure}[b]{0.32\textwidth}
        \centering
        \includegraphics[width=\textwidth,height=0.22\textheight]{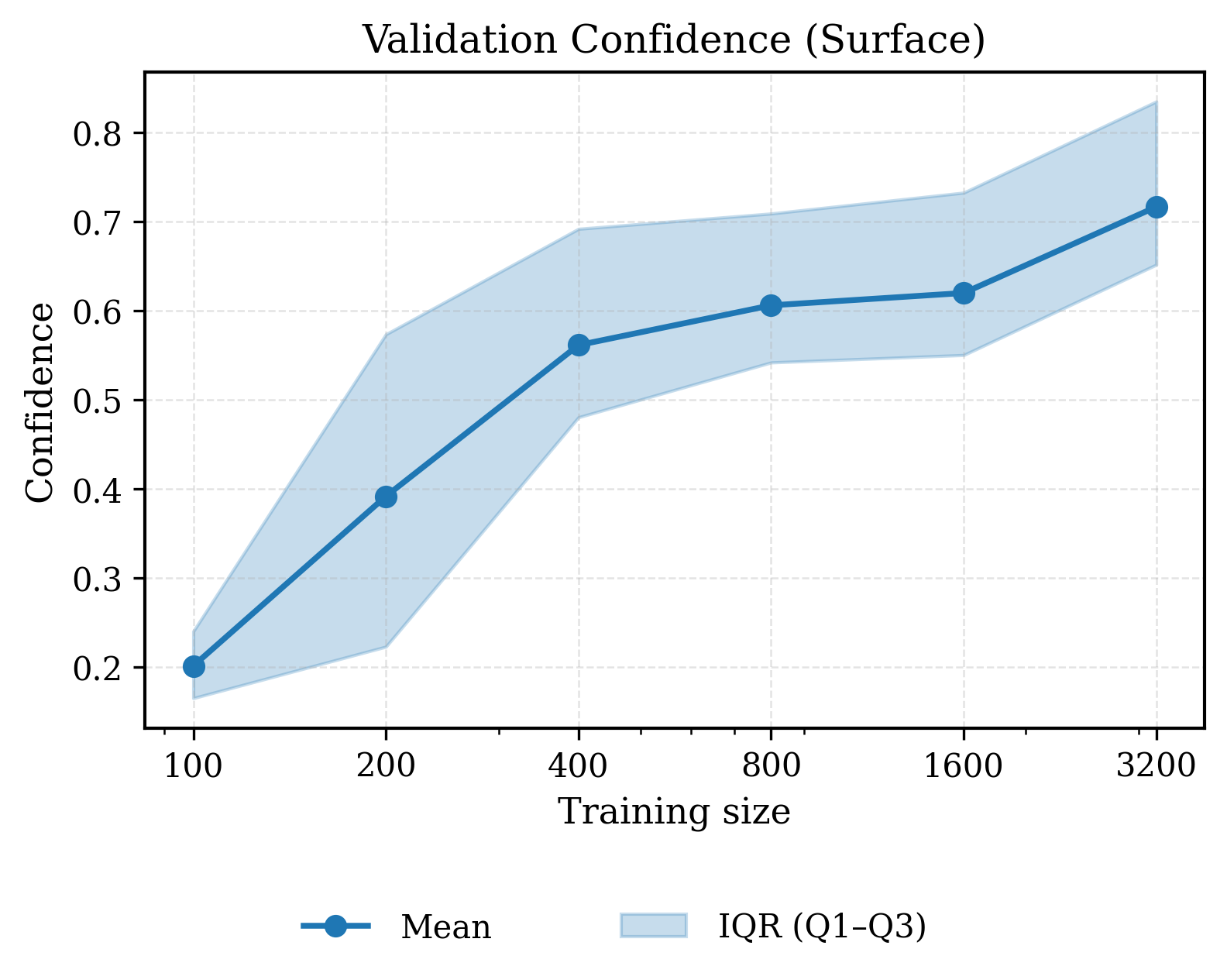}
        \caption{Confidence for Surface Estimation}
        \label{fig:confidence_surface_iqr}
    \end{subfigure}

    \caption{\textbf{Validation results across different training sizes.} (a) shows MAPE distributions for volume and surface, while (b) and (c) present confidence scores for the volume estimation and the surface estimation, respectively.}
    \label{fig:validation_combined}
\end{figure*}

This analysis evaluates how varying the size of the training dataset affects the MAPE for volume and surface area estimations, as well as predicted confidence scores. We trained the model using datasets of increasing sizes: 100, 200, 400, 800, 1600, and 3200 corals, with each coral represented by 30 input images. All models were evaluated on a fixed test dataset comprising 800 corals to ensure consistent comparisons. \Figure \ref{fig:validation_mape_confidence} shows the mean MAPE over the tested corals for both the volume and the surface across these training sizes. It also depicts the Inter-Quartile Range (IQR) of the MAPE on the same testing dataset. For both metrics, the MAPE decreases with increasing training sizes. For surface area estimation (blue curve), MAPE is initially high at $32.35\%$ with 100 samples, but it drops sharply to below $23.24\%$ at 200 samples and continues to decline gradually to $10.07\%$ at 3200 samples. In contrast, volume estimation (orange curve) starts at a lower MAPE of $26.24\%$ with 100 samples and decreases more steadily, reaching $12.40\%$ at 3200 samples. Additionally, the IQR bands for MAPE narrow as the training size increases, indicating reduced variability in performance across runs. These trends highlight that both tasks benefit substantially from larger datasets, with the most rapid improvements occurring early (between 100 and 200 samples) and diminishing thereafter. For a deeper understanding of the MAPE distribution across samples for both volume and surface predictions, detailed histograms are provided in the supplementary material.

Figure \ref{fig:confidence_volume_iqr} and Figure \ref{fig:confidence_surface_iqr} illustrate the evolution of predicted confidence scores for volume and surface area estimations, respectively.  The mean confidence scores increase steadily with training size for both metrics, with the most substantial gains observed between 100 and 400 samples, after which the improvements slow. The IQR remains relatively constant for training sizes of 400 samples and larger, suggesting stable variability in confidence predictions at higher dataset sizes.

\subsection{Evaluation of Feature Map Extraction Module}
In this subsection, we evaluate several parameters related to the VGGT-based module that extracts the point clouds from the input images. Our analysis focuses on 3 key factors: the number and spatial distribution of input images, the resolution of point clouds processed by the decoders, and the impact of data augmentation during training.
We evaluated the framework using a DGCNN backbone for volume regression from point clouds of coral structures. This experiment does not tackle the surface regression task. The network was trained with two standard regression losses: L1 (Mean Absolute Error, MAE) and Mean Squared Error (MSE).

Table~\ref{tab:80degree} reports the results for the limited-view experiments, where we used 16 input images within an $80$-degree spherical section (spanning $80^{\circ}$ horizontally and vertically). For these experiments, we tested two point cloud resolutions ($2048$ points and approximately $22,000$ points) and compared using all points versus only the top $25\%$ ranked by VGGT-predicted confidence. The results indicate that higher resolution and using all points generally improved performance, with the best outcome achieved using L1 loss, $22,000$ points, and all points (MAE: $2.37\cdot 10^{-2}$, RMSE: $3.80\cdot 10^{-2}$). Although MSE loss performs slightly worse than the L1 loss, it exhibits the same tendency for point cloud resolution and selection strategy.

\begin{table}[h]
\centering
\scriptsize  
\caption{Ablation study results for limited-view experiments: Varying loss type, point cloud resolution, and points selection. The \textbf{best} and \underline{second-best} scores are highlighted.}
\label{tab:80degree}
\begin{tabular}{lcccccc}
\toprule
\textbf{Loss Type} & \textbf{Resolution} & \textbf{Selection} & \textbf{MAE ↓} & \textbf{RMSE ↓} \\
\midrule
L1 & 2048 & top 25\% & 2.62e-2 & 4.33e-2 \\
L1 & 2048 & all points & 2.55e-2 & 4.14e-2 \\
L1 & 22k & top 25\% & 2.58e-2 & 4.51e-2 \\
L1 & 22k & all points & \textbf{2.37e-2} & \textbf{3.80e-2} \\
MSE & 2048 & top 25\% & 2.79e-2 & 4.58e-2 \\
MSE & 2048 & all points & 2.66e-2 & 4.29e-2 \\
MSE & 22k & top 25\% & 2.93e-2 & 4.59e-2 \\
MSE & 22k & all points & \underline{2.52e-2} & \underline{3.93e-2} \\
\bottomrule
\end{tabular}
\end{table}

In the multi-view experiments, we captured $30$ images along a horizontal trajectory on the sphere, providing full $360^{\circ}$ horizontal coverage. Table~\ref{tab:allangles} summarizes these results, including the incorporation of data augmentation, where a random subset of points is used for each coral at every epoch. The results demonstrate that data augmentation yields the best performance and reduces the impact of point cloud resolution, as all points contribute to the training process after a large number of epochs.

\begin{table}[h]
\centering
\scriptsize  
\caption{Ablation study results for multi-view experiments: Varying loss type, point cloud resolution, and data augmentation. The \textbf{best} and \underline{second-best} scores are highlighted.}
\label{tab:allangles}
\begin{tabular}{lcccc}
\toprule
\textbf{Loss Type} & \textbf{Resolution} & \textbf{Augmentation} & \textbf{MAE ↓} & \textbf{RMSE ↓} \\
\midrule
L1 & 2048 & Augmented & 1.69e-2 & \textbf{2.22e-2} \\
L1 & 2048 & Non Augmented & 1.63e-2 & 3.12e-2 \\
L1 & 22k & Augmented & \underline{1.44e-2} & \underline{2.76e-2} \\
L1 & 22k & Non Augmented & 1.75e-2 & 3.37e-2 \\
MSE & 2048 & Augmented & 1.77e-2 & 3.21e-2 \\
MSE & 2048 & Non Augmented & 1.65e-2 & 2.94e-2 \\
MSE & 22k & Augmented & \textbf{1.37e-2} & 2.77e-2 \\
MSE & 22k & Non Augmented & 2.17e-2 & 3.69e-2 \\
\bottomrule
\end{tabular}
\end{table}

\subsection{Comparison of Decoder Architectures}
\label{ssec:architecture_analysis}
To identify the most effective decoder for processing the extracted 4D feature map and predicting coral volume and surface area, we evaluated several architectures: standard Graph Convolutional Network (GCN)~\cite{kipf2017semisupervisedclassificationgraphconvolutional}, Graph Isomorphism Network (GIN)~\cite{xu2018powerful},  Graph Sample and Aggregate (GraphSAGE)~\cite{hamilton2018inductiverepresentationlearninglarge}, PointTransformer~\cite{zhao2021pointtransformer}, PointMLP~\cite{ma2022rethinkingnetworkdesignlocal}, and Dynamic Graph CNN (DGCNN)~\cite{wang2019dynamicgraphcnnlearning}.

In these experiments, all models were trained on random subsets of $22,000$ points from each point cloud, with a new subset sampled per training epoch. This strategy helps manage GPU memory usage, especially for architectures like DGCNN, whose memory requirements grow with the number of points.

As detailed in Table~\ref{tab:architectures_surf_vol}, DGCNN demonstrated the strongest overall performance, achieving the lowest or second-lowest metrics across most evaluation measures (MAE, MAPE, and RMSE) for both volume and surface area predictions. This consistent superiority stems from DGCNN's dynamic edge feature learning, which adeptly captures the local geometric relationships in the irregular and complex shapes of coral structures. Consequently, we selected DGCNN as the decoder architecture in our final framework for both volume and surface area regression tasks.

\begin{table}[!h]
\centering
\renewcommand{\arraystretch}{1.15}  
\scriptsize
\caption{Comparison of different decoder architectures for volume and surface prediction. The \textbf{best} and \underline{second-best} scores are highlighted.}
\label{tab:architectures_surf_vol}
\resizebox{1.01\columnwidth}{!}{  
\begin{tabular}{lcccc}
\toprule
\textbf{Architecture} & \textbf{MAE ↓} & \makecell{\textbf{MAPE ↓} \\ (mean)} & \makecell{\textbf{MAPE ↓} \\ (median)} & \textbf{RMSE ↓} \\
\midrule
GCN & \underline{\Volume{1.61e-2}} / \Surface{8.17e-1} & \Volume{22.00} / \Surface{14.98} & \Volume{13.02} / \Surface{11.22} & \underline{\Volume{2.70e-2}} / \Surface{1.22} \\
GIN & \Volume{1.81e-2} / \Surface{\textbf{4.88e-1}} & \Volume{25.05} / \Surface{13.64} & \Volume{14.43} / \underline{\Surface{9.98}} & \Volume{2.97e-2} / \underline{\Surface{1.10}} \\
GraphSAGE & \Volume{1.75e-2} / \Surface{7.61e-1} & \Volume{22.08} / \Surface{\textbf{13.07}} & \Volume{13.55} / \Surface{10.22} & \Volume{3.01e-2} / \Surface{1.18} \\
PointTransformer & \Volume{1.71e-2} / \Surface{8.04e-1} & \underline{\Volume{21.21}} / \Surface{13.81} & \underline{\Volume{12.05}} / \Surface{11.65} & \Volume{3.05e-2} / \Surface{1.20} \\
PointMLP & \Volume{1.96e-2} / \Surface{7.97e-1} & \Volume{28.28} / \Surface{14.000} & \Volume{14.03} / \Surface{\textbf{9.45}} & \Volume{3.58e-2} / \Surface{1.25} \\
DGCNN & \Volume{\textbf{1.40e-2}} / \underline{\Surface{6.90e-1}} & \Volume{\textbf{20.23}} / \underline{\Surface{13.57}} & \Volume{\textbf{11.42}} / \Surface{10.02} & \Volume{\textbf{2.02e-2}} / \Surface{\textbf{0.95}} \\
\bottomrule
\end{tabular}
}
\end{table}

\subsection{Evaluation of Different Loss Weights}
To assess the impact of loss hyperparameters on the performance of our volume and surface prediction framework, we evaluated six variants of the proposed hybrid loss function. By systematically varying the weights, we analyzed how these configurations affect regression accuracy metrics: the MAE, the MAPE (mean and median), and the RMSE. The loss configurations are summarized in Table~\ref{tab:losses}, which details the values of 
$\alpha$, $\beta$, $\delta$, $\gamma$ in equation~\ref{eq:total_loss_eq} for each variant. These variations aim to explore how emphasizing different loss components influences the model's performance.

\begin{table}[h!]
\centering
\caption{Loss configurations with different weight values.}
\scriptsize
\label{tab:losses}
\begin{tabular}{lcccccc}
\toprule
\textbf{Params} & \textbf{Loss 1} & \textbf{Loss 2} & \textbf{Loss 3} & \textbf{Loss 4} & \textbf{Loss 5} & \textbf{Loss 6} \\
\midrule
$\alpha$ & 0.50 & 0.50 & 0.50 & 0.50 & 0.30 & 0.30 \\
$\beta$ & 0.35 & 0.15 & 0.15 & 0.35 & 0.35 & 0 \\
$\delta$ & 0.20 & 0 & 0.20 & 0 & 0.20 & 0 \\
$\gamma$ & 0.10 & 0.10 & 0.10 & 0.10 & 0.10 & 0.10 \\
\bottomrule
\end{tabular}
\end{table}

Table~\ref{tab:losses_surf} compares the performance of these loss variants for both volume and surface predictions. From the obtained results, \textbf{Loss 5} outperforms others for volume prediction, achieving the lowest MAE, MAPE (mean), and RMSE, and a second-lowest MAPE (median). This suggests that using balanced weights for the different loss components is the most effective strategy for volume estimation. This may be attributed to the non-linear distribution of volume values, where absolute metrics like MAE alone are less informative. Instead, relative loss accounts for proportional errors. Moreover, $\operatorname{NLL}_{\log}$ enhances accuracy by addressing the logarithmic scale of the data. This term' impact is mainly visible when we compare Loss 5 to Loss 1.

In contrast, for the surface prediction task, \textbf{Loss 6} yields the best results across most metrics (except RMSE), indicating that allocating more weight to relative loss alone improves performance. This difference highlights how the optimal loss configuration depends on the prediction type: volume benefits from a balanced weighting approach, while surface predictions are more sensitive to relative error weighting.

\begin{table}[!h]
\centering
\renewcommand{\arraystretch}{1.15}  
\scriptsize
\caption{Combined comparisons of volume and surface results for different loss hyperparameters (format: \Volume{Volume} / \Surface{Surface}). The \textbf{best} and \underline{second-best} scores are highlighted.}
\label{tab:losses_surf}
\resizebox{0.93\columnwidth}{!}{  
\begin{tabular}{lcccc}
\toprule
\textbf{Losses} & \textbf{MAE ↓} & \makecell{\textbf{MAPE ↓} \\ (mean)} & \makecell{\textbf{MAPE ↓} \\ (median)} & \textbf{RMSE ↓} \\
\midrule
Loss 1 & \underline{\Volume{1.50e-2}} / \underline{\Surface{6.47e-1}}  & \underline{\Volume{16.01}} / \underline{\Surface{11.51}} & \Volume{\textbf{8.93}} / \Surface{9.38} & \underline{\Volume{2.51e-2}} / \Surface{\textbf{1.03}} \\
Loss 2 & \Volume{1.78e-2} / \Surface{7.17e-1}  & \Volume{18.44} / \Surface{12.20} & \Volume{12.47} / \Surface{9.59} & \Volume{3.01e-2} / \Surface{1.13} \\
Loss 3 & \Volume{1.71e-2} / \Surface{6.71e-1}  & \Volume{17.80} / \Surface{11.95} & \Volume{11.30} / \Surface{9.41} & \Volume{2.85e-2} / \Surface{1.06} \\
Loss 4 & \Volume{1.75e-2} / \Surface{6.59e-1}  & \Volume{17.56} / \Surface{11.64} & \Volume{12.30} / \underline{\Surface{8.81}} & \Volume{2.97e-2} / \Surface{1.08} \\
Loss 5 & \Volume{\textbf{1.41e-2}} / \Surface{6.82e-1}  & \Volume{\textbf{15.59}} / \Surface{12.43} & \underline{\Volume{10.52}} / \Surface{9.85} & \Volume{\textbf{2.40e-2}} / \Surface{1.07} \\
Loss 6 & \Volume{1.54e-2} / \Surface{\textbf{6.25e-2}}  & \Volume{16.13} / \Surface{\textbf{10.85}} & \Volume{10.66} / \Surface{\textbf{8.49}} & \Volume{2.54e-2} / \underline{\Surface{1.04}} \\
\bottomrule
\end{tabular}
}
\end{table}

\section{Conclusion}
\label{sec:conclusion}

This paper presents an efficient and generalizable approach for estimating the volume and surface area of corals using only 2D multi-view images. Instead of relying on dense 3D supervision or high-end scanning systems, we adopt a lightweight pipeline that reconstructs 3D point clouds through VGGT and infers geometric properties via a DGCNN-based decoder.

The proposed framework is designed to be robust to challenges posed by the complex and variable coral morphologies. To handle prediction uncertainty and scale variance, we formulated a hybrid loss function that balances probabilistic and deterministic terms, along with absolute and relative error metrics.

Despite training with a modest dataset, our approach demonstrates strong performance across diverse shapes and scales, enabling fast and accurate coral morphology analysis without reliance on costly hardware or extensive annotations. This paves the way for practical applications in large-scale coral monitoring and growth estimation, supporting conservation efforts in underwater environments.\\

In future work, we will enhance the applicability and accuracy of our framework by capturing real-world underwater images of corals, which can be used for fine-tuning the model on authentic data and validating its performance in real-world scenarios. This step is essential to address potential domain gaps between synthetic and actual corals, improving generalization and reliability. We also aim to develop an automatic method for determining the appropriate scale factor to convert normalized volume and surface measurements to absolute values.



{
    \small
    \bibliographystyle{ieeenat_fullname}
    \bibliography{biblio}
}
\clearpage
\setcounter{page}{1}
\maketitlesupplementary

\section{Additional implementation details}

\subsection{Dataset}
The 3D assets originate from \textit{Infinigen} were first exported in \texttt{.blend} format, then converted to \texttt{.obj}. We processed each \texttt{.obj} with \textit{ManifoldPlus} to obtain watertight (2-manifold) meshes, a prerequisite for accurate geometric ground-truth computations of volume and surface area. The watertight meshes were then loaded with PyVista and Trimesh: PyVista facilitated off-screen rendering (to produce the multi-view RGB images used as input for our pipeline), 3D visualization, and the extraction of mesh bounds and normals; Trimesh handled geometric repairs (e.g., fixing normals, removing degenerate faces) and, together with PyVista, was used to compute ground-truth metrics (volume and surface area). From these renderings and meshes we precomputed point features stored in \texttt{.pt} files, each containing an array of $N \times 4$ points and the corresponding ground-truth volume. We split the dataset into training and validation subsets using fixed indices, resulting in 1170 training objects and 291 validation objects. Figure~fig:manycorals shows some examples of the generated corals that we used in our training process. 

\subsection{Decoder Model Architecture.}
The decoder model is based on a Dynamic Graph Convolutional Neural Network (DGCNN) architecture. We used $k = 30$ nearest neighbors to construct the local graph features. The encoder consists of five EdgeConv blocks with output channels $\{64, 64, 128, 256, 256\}$, each followed by instance normalization and LeakyReLU activation. The extracted features are globally aggregated via adaptive max pooling. The final representation is processed through a multi-layer perceptron with hidden dimensions $\{512, 128\}$ and dropout ($p=0.3$), followed by a fully connected layer producing two outputs: the predicted volume or surface area $\mu_{\text{real}}$ and a raw uncertainty estimate $\log \sigma^2$.

\subsection{Training Details.}
We trained our model to predict the real-valued volume and surface area of coral structures from point cloud features extracted from multi-view renderings. The input data consisted of point clouds with per-point 4D features $(x, y, z, f)$, where $f$ is an image-based descriptor aggregated from multi-view geometry. Each object was subsampled to a fixed number of 40000 points per sample.
We trained the model using the AdamW optimizer with a learning rate of $1.6 \times 10^{-4}$ and weight decay of $1.0 \times 10^{-4}$. The learning rate was scheduled using Cosine Annealing with a minimum learning rate $\eta_{\text{min}} = 2 \times 10^{-6}$ and a cycle length of $T_{\text{max}} = 60$ epochs. Training was performed for a maximum of 650 epochs with early stopping after 35 epochs without improvement in validation loss. The batch size was set to 1 due to memory constraints and varying point cloud sizes.
The model achieving the lowest validation loss was saved as the final checkpoint. At the end of training, we exported all metrics and predictions to \texttt{.csv} files.
On average, each epoch required approximately 14 minutes and 55 seconds for training and 3 minutes and 12 seconds for validation, for a total of about 18 minutes per epoch. When considering a larger number of training samples, checkpoints obtained after training the models on less data were utilized to save time.

\section{Faulty watertight reconstruction}

\begin{figure}[ht!]
    \centering
    \begin{subfigure}[b]{0.27\textwidth}
        \centering
        \includegraphics[width=\textwidth]{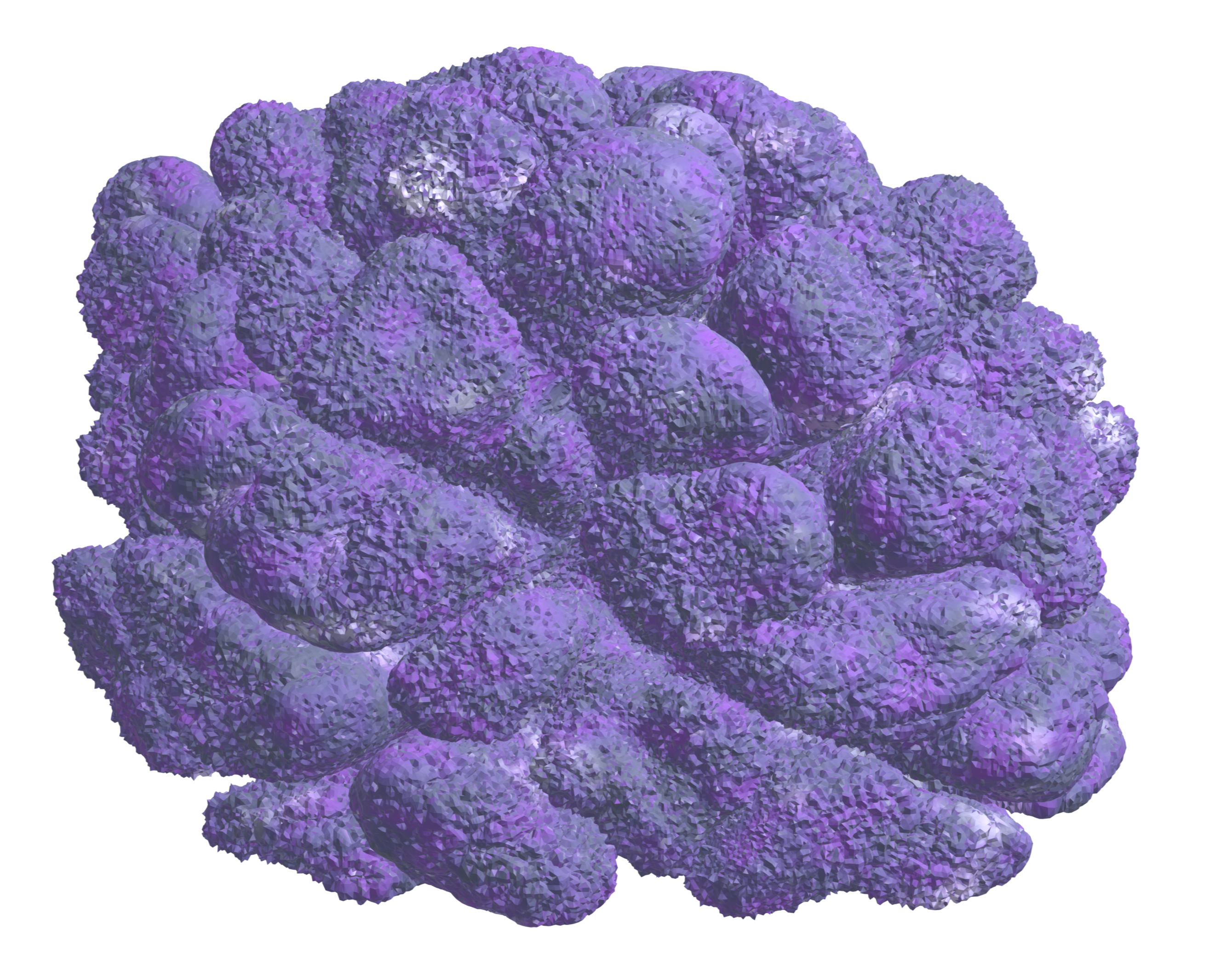}
        \caption{}
        \label{fig:image1}
    \end{subfigure}
    \hfill
    \begin{subfigure}[b]{0.375\textwidth}
        \centering
        \raisebox{0mm}{\hspace{9.8mm}%
            \includegraphics[width=\textwidth]{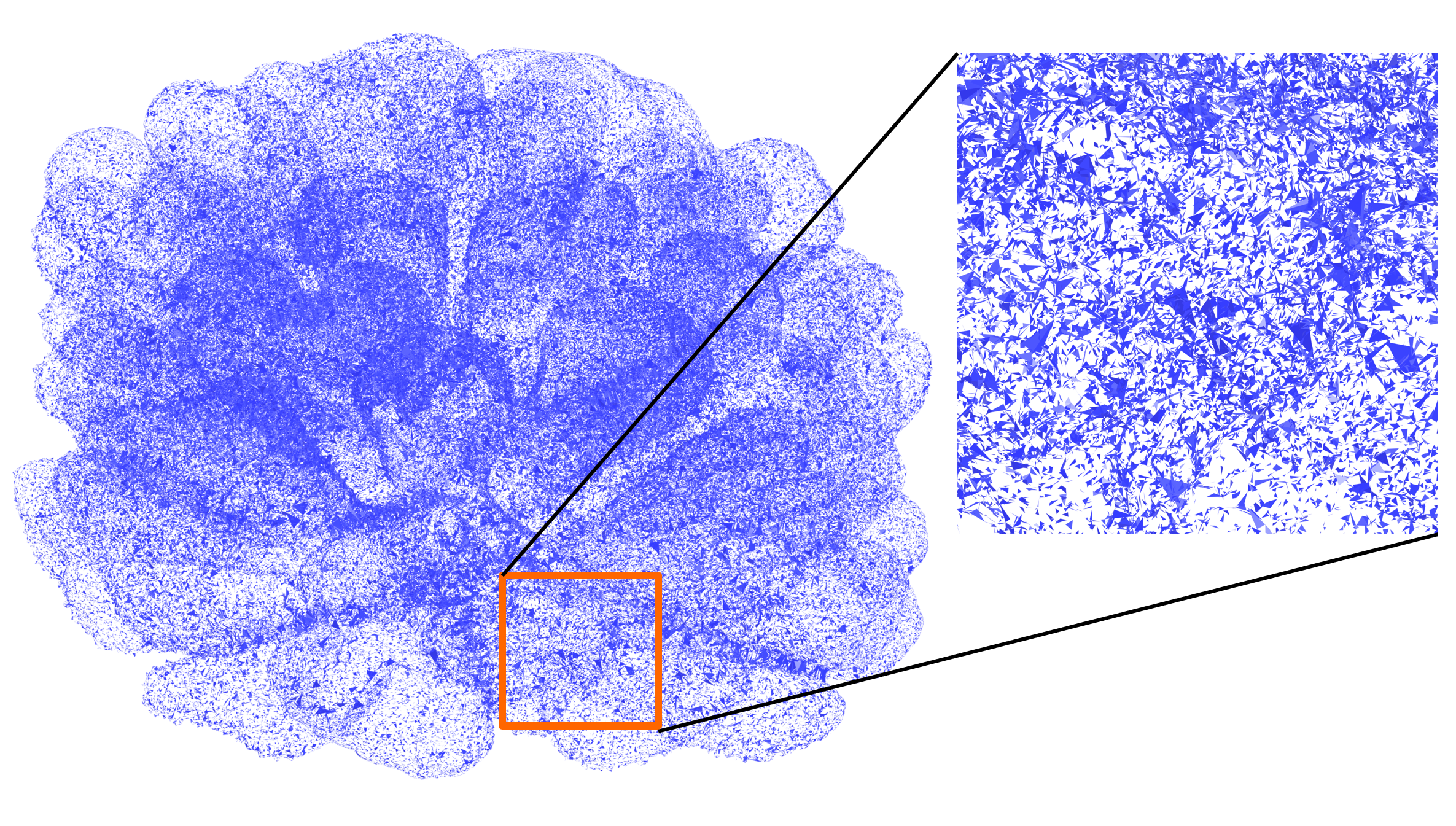}
        }
        \caption{}
        \label{fig:image2}
    \end{subfigure}

    \caption{\textbf{Example of a faulty watertight reconstruction.} Original non-watertight reconstruction (a), watertight version produced by ManifoldPlus (b), which exhibits severe artifacts like thin sheets and visible stretched triangles due to the watertightness process. These errors can drastically distort the estimated volume, invalidating the ground-truth value.}
    \label{fig:twoimages}
\end{figure}

To compute the volume and surface area of a mesh, we need to make it first watertight using \textit{ManifoldPlus}. The synthetic meshes generated by \textit{Infinigen} usually present severe artifacts like thin sheets and visible stretched triangles as shown in Figure~\ref{fig:twoimages}.

\begin{figure*}[!ht]
    \centering
    \includegraphics[width=0.85\textwidth]{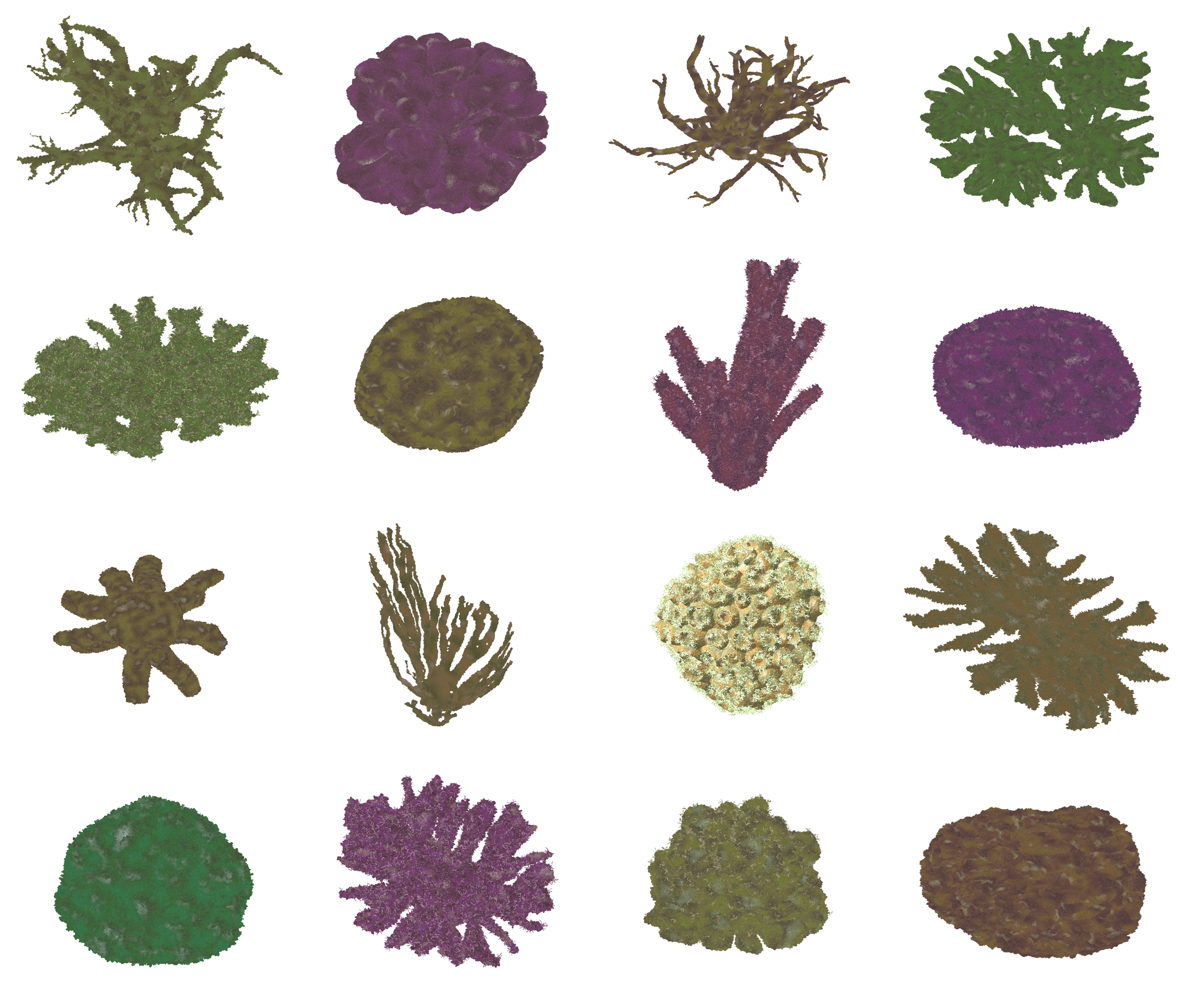}
    \caption{Example coral structures from our dataset. These diverse coral morphologies were used to train and evaluate the proposed model.}
    \label{fig:manycorals}
\end{figure*}


\section{Additional results}

\subsection{Comparison with Trellis}
This subsection presents supplementary results from our evaluation of the Trellis-based method against our approach for predicting coral volume and surface area, as shown in Tables~\ref{tab:metrics_compact} and~\ref{tab:volume_surface_results}. Table~\ref{tab:metrics_compact} provides additional performance metrics, including MAPE, MAE, and RMSE, which were not all covered in the main paper. Table~\ref{tab:volume_surface_results} displays the detailed predictions for each of the 15 corals in the comparison, demonstrating that our method outperforms Trellis in nearly all cases. Furthermore, it highlights the substantial variability in coral sizes, with volumes spanning a factor of 67 between the smallest and largest, and surface areas varying by a factor of 10.

\begin{table*}[h!]
\centering
\caption{Comparison of average errors (MAPE, MAE, RMSE) between our method and Trellis for Volume and Surface.}
\label{tab:metrics_compact}
\scriptsize
\begin{tabular}{l l | c c c}
\toprule
 & Model & \textbf{MAPE $\downarrow$} & \textbf{MAE $\downarrow$} & \textbf{RMSE $\downarrow$} \\
\midrule
\multirow{2}{*}{\textbf{Volume}} 
  & Trellis  & 60.81\%          & 0.0570          & 0.0820 \\
  & \textbf{Ours}     & \textbf{10.26\%} & \textbf{0.0045} & \textbf{0.0067} \\
\midrule
\multirow{2}{*}{\textbf{Surface}} 
  & Trellis  & 55.41\%          & 3.195           & 4.312 \\
  & \textbf{Ours}     & \textbf{7.56\%}  & \textbf{0.210}  & \textbf{0.574} \\
\bottomrule
\end{tabular}
\end{table*}

\begin{table*}[h!]
\centering
\caption{Prediction results for Volume and Surface with Relative Error comparison for both our model and Trellis.}
\label{tab:volume_surface_results}
\resizebox{0.95\textwidth}{!}{%
\begin{tabular}{c | c c c c c | c c c c c}
\toprule
\multirow{2}{*}{\textbf{Coral ID}} & \multicolumn{5}{c|}{\textbf{Volume}} & \multicolumn{5}{c}{\textbf{Surface}} \\
\cmidrule(lr){2-6} \cmidrule(lr){7-11}
 & \textbf{GT} & \textbf{Pred. (ours)} & \textbf{Pred. (Trellis)} & \textbf{Rel. Error (ours) $\downarrow$} & \textbf{Rel. Error (Trellis) $\downarrow$} 
 & \textbf{GT} & \textbf{Pred. (ours)} & \textbf{Pred. (Trellis)} & \textbf{Rel. Error (ours) $\downarrow$} & \textbf{Rel. Error (Trellis) $\downarrow$} \\
\midrule
1  & 0.03445 & 0.02753 & 0.10615 & \textbf{20.09} & 54.74 & 3.75060 & 3.65185 & 4.16506 & \textbf{2.63} & 3.16 \\
2  & 0.12652 & 0.13494 & 0.01837 & \textbf{6.65}  & 18.46 & 9.50442 & 12.2495 & 1.40335 & \textbf{28.88} & 58.67 \\
3  & 0.01551 & 0.02014 & 0.03654 & \textbf{29.88} & 83.42 & 3.39523 & 3.63119 & 2.17491 & \textbf{6.95}  & 53.93 \\
4  & 0.22040 & 0.23807 & 0.03531 & \textbf{8.02}  & 54.10 & 4.72044 & 5.01023 & 2.44817 & \textbf{6.14}  & 46.48 \\
5  & 0.19560 & 0.18763 & 0.00432 & \textbf{4.07}  & 33.35 & 8.81682 & 8.82157 & 1.06862 & \textbf{0.05}  & 41.04 \\
6  & 0.01499 & 0.01466 & 0.00286 & \textbf{2.20}  & 48.47 & 5.34308 & 5.51655 & 0.97420 & \textbf{3.25}  & 60.90 \\
7  & 0.13650 & 0.13797 & 0.11802 & \textbf{1.08}  & 46.33 & 9.80091 & 8.48931 & 4.50957 & \textbf{13.38} & 60.22 \\
8  & 0.00329 & 0.00396 & 0.03961 & 20.51 & \textbf{5.39}  & 1.41976 & 1.63870 & 2.39011 & \textbf{15.42} & 67.19 \\
9  & 0.01446 & 0.01340 & 0.05332 & \textbf{7.30}  & 96.90 & 2.42110 & 1.99456 & 2.70213 & \textbf{17.62} & 63.29 \\
10 & 0.12604 & 0.11488 & 0.07260 & \textbf{8.85}  & 74.07 & 11.9187 & 11.4698 & 3.03870 & \textbf{3.77}  & 60.57 \\
11 & 0.07693 & 0.07624 & 0.05636 & \textbf{0.89}  & 119.60 & 4.57429 & 4.67689 & 2.53935 & \textbf{2.24}  & 69.20 \\
12 & 0.00849 & 0.00779 & 0.00982 & \textbf{8.24}  & 92.24 & 1.17439 & 1.27465 & 1.16797 & \textbf{8.54}  & 87.71 \\
13 & 0.01007 & 0.00975 & 0.00729 & \textbf{3.22}  & 27.60 & 1.96958 & 1.91217 & 0.98592 & \textbf{2.92}  & 49.94 \\
14 & 0.01578 & 0.01109 & 0.03394 & \textbf{29.73} & 86.21 & 5.17128 & 5.10079 & 2.34874 & \textbf{1.36}  & 70.31 \\
15 & 0.00839 & 0.00866 & 0.08444 & \textbf{3.23}  & 71.30 & 1.65140 & 1.64821 & 4.18178 & \textbf{0.19}  & 38.53 \\
\bottomrule
\end{tabular}
}
\end{table*}

\subsection{Validation Relative Percentage Error Distribution Analysis}
\begin{figure*}[h!]
    \centering
    \includegraphics[width=0.85\textwidth]{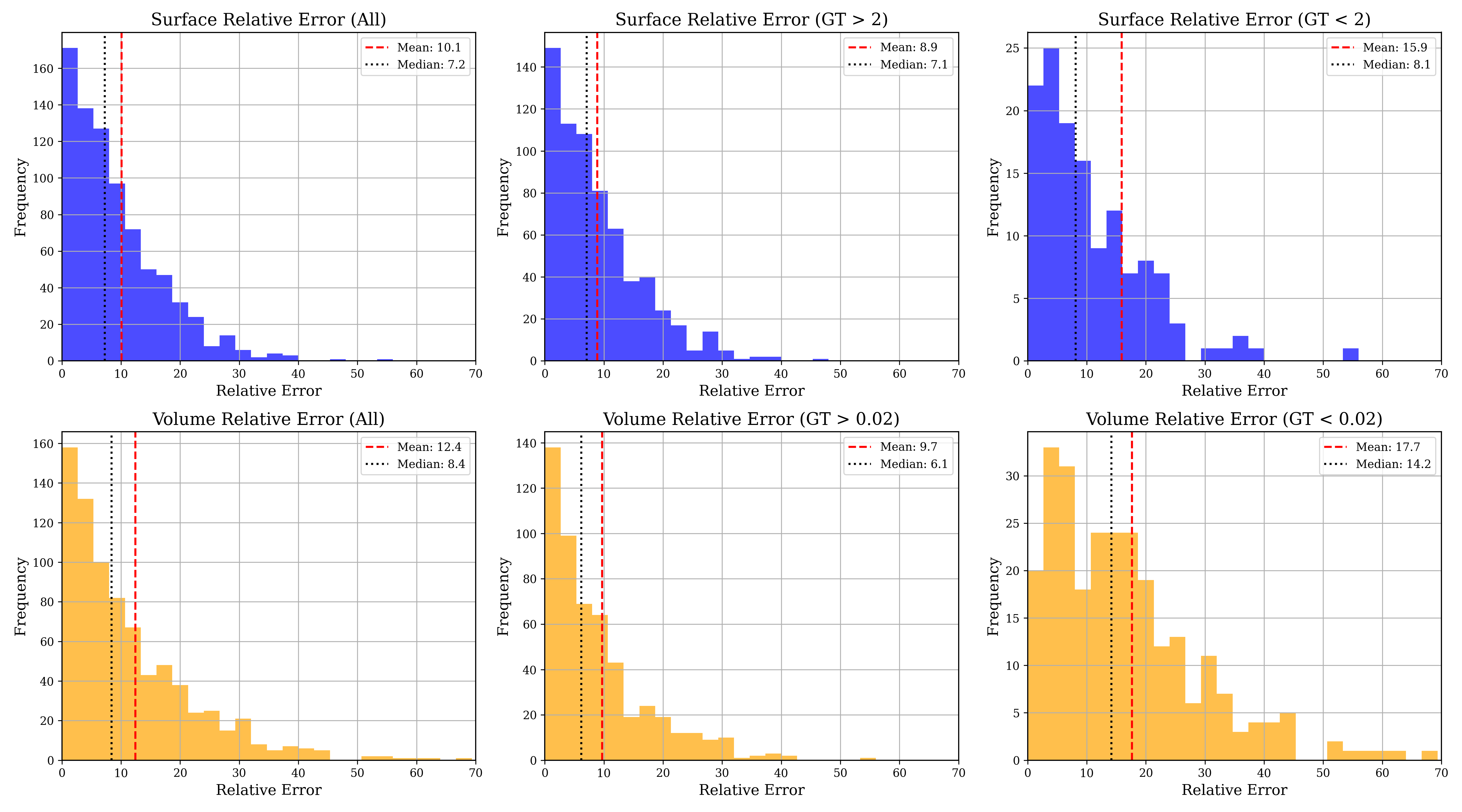}
    \caption{Histograms of the Relative Error for surface (top row, blue) and volume (bottom row, orange) predictions across all validation samples. The leftmost plots show the overall distribution, while the center and right plots break down the results based on the ground truth (GT) magnitude: GT > 2 for surface and GT > 0.02 for volume (center), and GT < 2 or < 0.02 respectively (right). We observe that when filtering out small ground truth values (center), the distributions shift slightly to the left, indicating lower errors on larger structures. Conversely, for samples with low GT values (right), the distributions are clearly shifted to the right, reflecting higher relative error percentages. Vertical dashed lines denote the mean (red) and median (black) values.}
    \label{fig:histo}
\end{figure*}

Figure~\ref{fig:histo} illustrates the distribution of Relative Error for predictions of coral surface area and volume across all validation samples. This visualization provides insights into the error patterns, particularly how prediction accuracy varies with the magnitude of the ground truth (GT) values.

The key observation from the histograms is a noticeable shift in the distributions when filtering by GT magnitude. For larger GT values (center column), the distributions shift slightly to the left, indicating lower relative errors and thus better predictive accuracy on bigger corals. Conversely, for smaller GT values (rightmost column), the distributions are shifted to the right, reflecting higher relative errors, which may be attributed to the increased sensitivity of percentage-based metrics to small relative errors in low-magnitude data. Vertical dashed lines in each histogram denote the mean (red) and median (black) relative errors, providing a quantitative summary of the central tendencies.

These patterns underscore the challenges in predicting properties of smaller corals and suggest potential avenues for model refinement, such as incorporating size-specific adjustments or alternative error metrics for low-magnitude cases.

\end{document}